\documentclass[10pt,twocolumn,letterpaper]{article}

\usepackage{cvpr}              

\usepackage{graphicx}
\usepackage{amsmath}
\usepackage{amssymb}
\usepackage{booktabs}
\usepackage{algorithm}
\usepackage{algorithmic}
\usepackage{extarrows}
\usepackage{multirow}

\usepackage[pagebackref,breaklinks,colorlinks]{hyperref}

\usepackage[capitalize]{cleveref}
\crefname{section}{Sec.}{Secs.}
\Crefname{section}{Section}{Sections}
\Crefname{table}{Table}{Tables}
\crefname{table}{Tab.}{Tabs.}

\def\cvprPaperID{5950} 
\def\confName{CVPR}
\def\confYear{2022}

\begin{document}


\newcommand*\samethanks[1][\value{footnote}]{\footnotemark[#1]}

\title{Siamese Contrastive Embedding Network for Compositional Zero-Shot Learning}

\author{Xiangyu Li, Xu Yang\thanks{Corresponding author.}, Kun Wei, Cheng Deng\samethanks, Muli Yang\\
School of Electronic Engineering, Xidian University, Xi'an 710071, China\\
{\tt\small xdu\_xyLi@stu.xidian.edu.cn, \{xuyang.xd, weikunsk, chdeng.xd, muliyang.xd\}@gmail.com}

}
\maketitle

\begin{abstract}
   Compositional Zero-Shot Learning (CZSL) aims to recognize unseen compositions formed from seen state and object during training. Since the same state may be various in the visual appearance while entangled with different objects, CZSL is still a challenging task. Some methods recognize state and object with two trained classifiers, ignoring the impact of the interaction between object and state; the other methods try to learn the joint representation of the state-object compositions, leading to the domain gap between seen and unseen composition sets. In this paper, we propose a novel Siamese Contrastive Embedding Network (SCEN)\footnote{Code: \url{https://github.com/XDUxyLi/SCEN-master}} for unseen composition recognition. Considering the entanglement between state and object, we embed the visual feature into a Siamese Contrastive Space to capture prototypes of them separately, alleviating the interaction between state and object. In addition, we design a State Transition Module (STM) to increase the diversity of training compositions, improving the robustness of the recognition model. Extensive experiments indicate that our method significantly outperforms the state-of-the-art approaches on three challenging benchmark datasets, including the recent proposed C-QGA dataset. 
\end{abstract}

\section{Introduction}
\label{sec:intro}

Humans possess the ability to compose their knowledge of known entities to generalize to novel concepts inherently. 
Given words, such as \emph{green horse}, people can combine the known state \emph{green} with the known object \emph{horse} immediately, although they have never seen the inexistent stuff.
To equip an AI system the similar ability, Compositional Zero-Shot Learning (CZSL)~\cite{misra2017red} is proposed, which aims to recognize unseen compositions composed of a set of seen states and objects.
In CZSL setting, each composition comprises two components, namely, state and object, where the compositions of train and test sets are disjoint.

\begin{figure}[t]
	\centering
	\includegraphics[width=1\columnwidth]{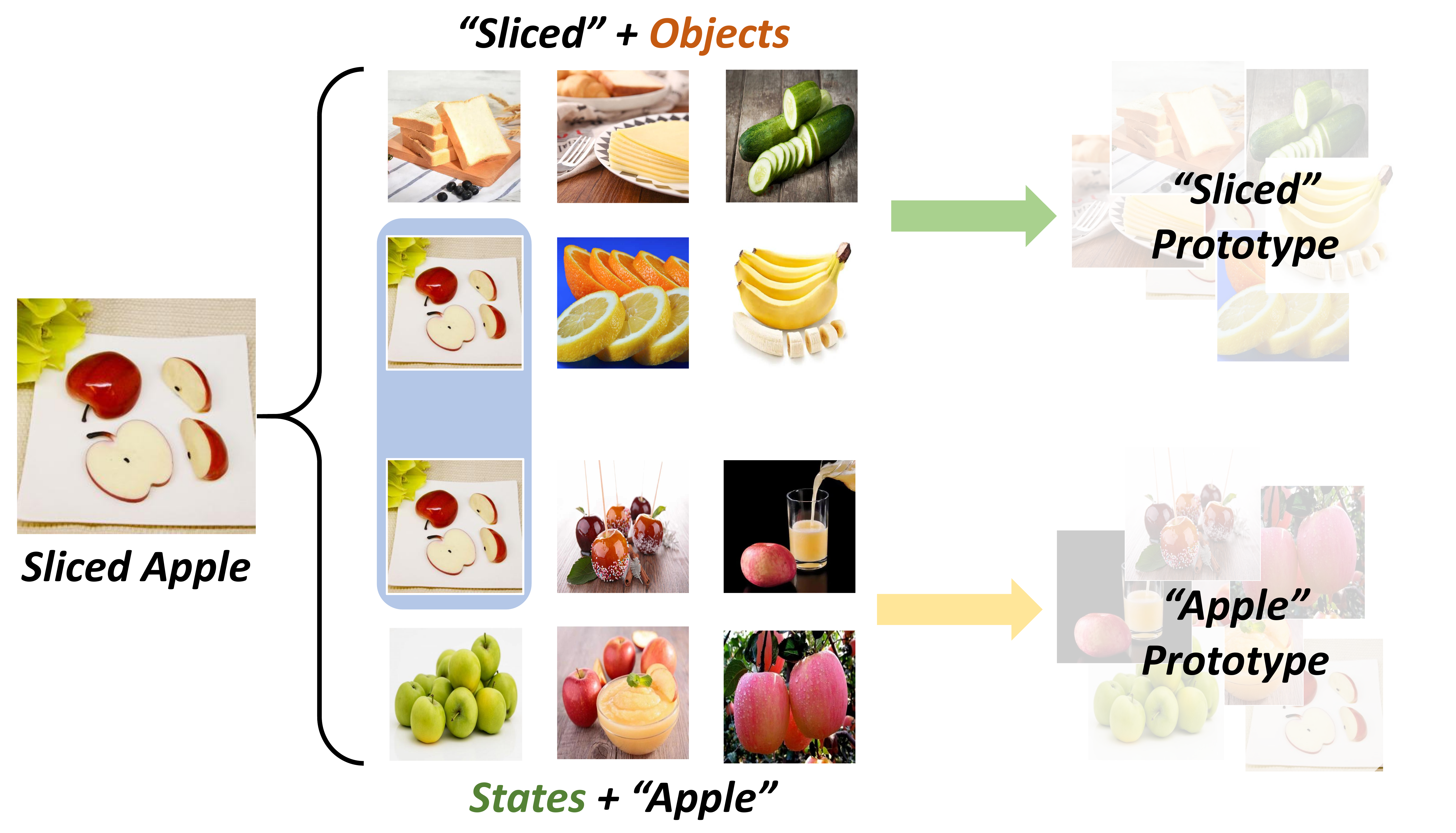} 
	\caption{The overall concept of our method. We aim to separately extract discriminative prototypes of state and object based on establishing state and object specific databases, which can generalize to represent corresponding properties.}
	\label{fig1}
\end{figure}

In order to infer unknown concepts such as \emph{green horse}, CZSL aims to understand the meaning of state \emph{green} and object \emph{horse} after trained on other compositional concepts that separately contain \emph{green} or \emph{horse}, e.g., \emph{green grasses} and \emph{young horse}.
The challenge of the task lies in the interaction degree between state and object that we cannot quantify, which gives rise to varying contextuality within different state-object combinations. 
For instance, we can not equate the state \emph{old} in \emph{old car} and that in \emph{old tiger}, since they are fundamentally distinct in visual presentations, which greatly hinders the recognition of novel compositions.

Existing mainstream methods~\cite{li2020symmetry, misra2017red, purushwalkam2019task} focus on converting such problem into a general supervised recognition task by training two classifiers for state and object, respectively. They aim to directly predict state and object from the original visual features, ignoring their entanglement.
Based on this problem, classifiers cannot capture discriminative state and object features, which potentially limits the recognition accuracy.
In addition, other methods~\cite{nan2019recognizing, nagarajan2018attributes} aim to learn a common embedding space where the compositions as well as visual features can be projected to narrow the distance between them, such as Euclidean distance. However, these methods, only regarding compositions as entities, neglect the domain gap between training and testing compositions, which can be simply confused by similar images from unseen compositions (e.g., \emph{young cat} and \emph{young tiger}).
Therefore, it is vital to excavate the discriminative prototypes of state and object to separate the interaction between them and consider the domain transfer between training and testing samples.

To address the problem mentioned above, we propose a Siamese Contrastive Embedding Network (SCEN) for recognizing novel compositions in this paper, aiming to excavate discriminative prototypes of state and object, respectively, as shown in~\cref{fig1}. 
To be specific, we first project the visual features into state/object-based contrastive spaces to gain the prototypes of state and object. Then, to excavate the discriminative prototypes by contrastive constraints, we set up specific databases named \emph{State-constant} and \emph{Object-constant} databases as positive samples. Besides, a shared irrelevant database is built up as a negative sample set, which is embedded into two contrastive spaces.
Benefiting from this learning paradigm, our proposed model can successfully excavate discriminative prototypes to represent the corresponding component. 
In addition, considering that the distribution between seen and unseen compositions is discrepant, we present a State Transition Module (STM), which generates the virtual but reasonable samples to augment the diversity of training data. In this way, the domain gap between seen and unseen composition sets can be mitigated effectively.


To sum up, our main contributions are as follows:

\begin{itemize}

\item We propose a novel Siamese Contrastive Embedding Network (SCEN) to excavate prototypes of state and object for successfully recognizing both seen and unseen compositions.
	
\item We present a State Transition Module (STM) to produce virtual samples and augment the diversity of training compositions, guiding the model to generalize to those compositions not existing in the training process, and alleviating the issue of model migration performance.
	
\item Comprehensive experimental results on three benchmark datasets demonstrate the effectiveness of our proposed approach, which outperforms the state-of-the-art CZSL methods.  
\end{itemize}

\begin{figure*}[t]
\centering
\includegraphics[width=0.9\textwidth]{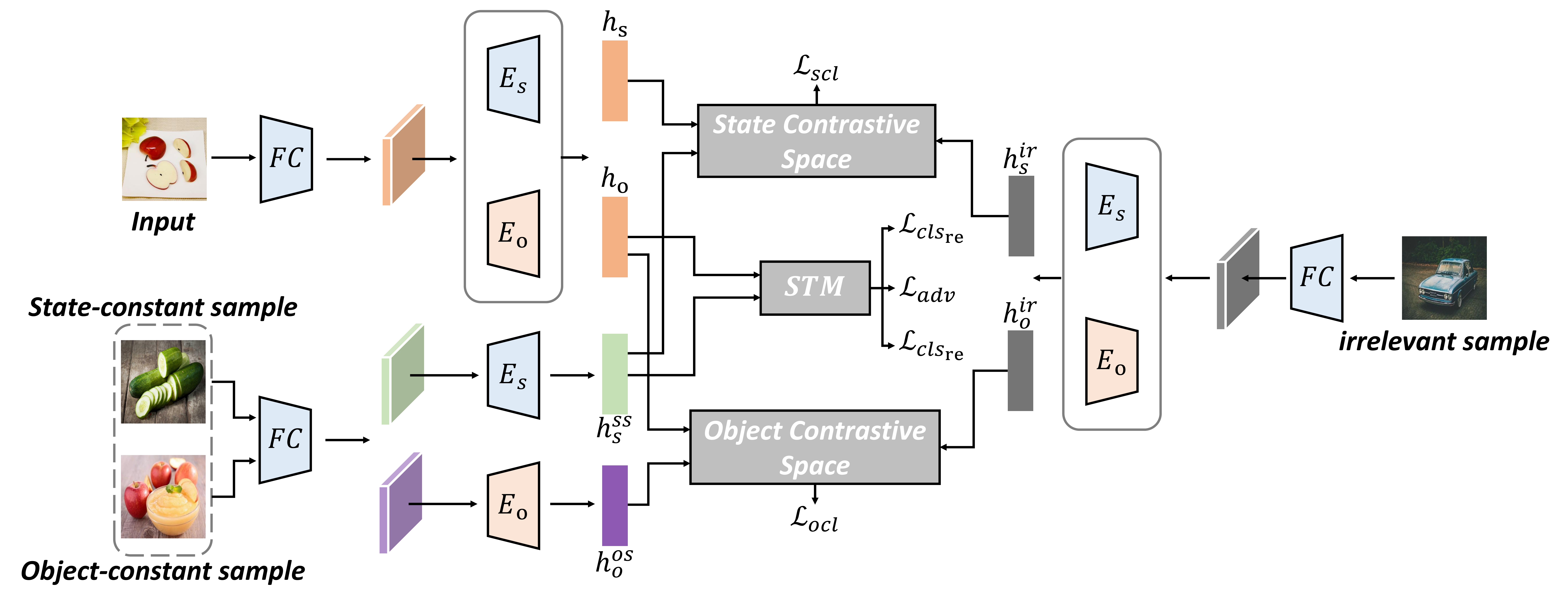} 
\caption{The framework of our SCEN method. The proposed method consists of State-Specific Encoder $ E_{s}$, Object-Specific Encoder $E_{o}$, and a State Transition Module (STM). The prototypes of state $h_{s}$ and object $h_{o}$ are separately encoded from Specific Encoders. Prototypes are trained conditioned on two different contrastive embedding spaces, where the anchor and negative samples are shared. In addition, STM aims to generate virtual samples with the help of the adversarial loss function, which diversifies and rationalizes the states of objects effectively. Along with the increasing of the realistic samples, Specific Encoders gradually gain the ability to extract discriminative prototypes that can be generalized to novel compositions.}
\label{fig_fr}
\end{figure*}

\section{Related Work}

\textbf{Compositional Zero-Shot Learning.} The goal of Compositional Zero-Shot Learning ~\cite{mikolov2013distributed, misra2017red, nagarajan2018attributes, li2020symmetry, purushwalkam2019task, gu2021class, wei2019adversarial} is to learn the compositionality of objects and their states from the training data and is tasked with the generalization to an unseen combination of these primitives. Compared with typical Zero-Shot Learning~\cite{lampert2013attribute, wei2020lifelong, li2021generalized} that utilizes inherent semantic descriptions or attributed vectors to recognize unseen instances, CZSL exploits transferable knowledge by two compositional parts as image labels: objects and states. There are two mainstream methods in this direction. The first mainstream approach is inspired by~\cite{biederman1987recognition, hoffman1984parts}, which learns a single classifier for recognition and a transformation module~\cite{misra2017red}. In addition,~\cite{nagarajan2018attributes} models each state as a linear transformation of objects.~\cite{yang2020learning} aims to learn disentangled and compositional primitives hierarchically. \cite{li2020symmetry} models objects to be symmetric under attribute transformations. Other methods try to learn the joint representation of the state-object compositions~\cite{atzmon2020causal, purushwalkam2019task, wang2019task}. They aim to learn a modular network to rewire the new compositions conditioned on each composition~\cite{purushwalkam2019task, wang2019task}. Recently, GCN~\cite{naeem2021learning} is proposed to utilize a causal graph to establish the relationship between state and object reasonably. However,these methods ignore the interaction between state and objects that brings a negative influence for compositions recognition.


As for~\cite{atzmon2020causal}, the author argues to tackle the CZSL problems through a causal graph where the latent features of primitives are independent from each other. However, it also neglects the discriminant analysis of state and object, which cannot excavate discriminative primitives for classification. In addition, there still exists a domain gap between seen and unseen compositions, although they are made up of the same states and objects, which potentially limits the performance of the model.

\textbf{Contrastive Learning.} Inspired by noise contrastive estimation~\cite{gutmann2010noise, mnih2013learning, zhao2021graph}, contrastive learning has attracted much attention which leads to major advances in self-supervised representation learning. An efficient way to get better contrastive learning is to employ large numbers of negative examples and design more semantically meaningful augmentations to create different view of images. SimCLR~\cite{chen2020simple} implements two data augmentation paths and a learnable non-linear transformation to train an encoder with a large batch by pulling the features embedding from the same images. Momentum Contrast (MoCo)~\cite{he2020momentum} is presented which enables building a large and consistent dictionary on-the-fly and transfers well to downstream tasks. Aiming at improving generalization in real domains, a contrastive synthetic-to-real generalization model~\cite{chen2021contrastive} is proposed to prevent overfitting to the synthetic domain by leveraging the pre-trained ImageNet knowledge. More recently, supervised contrastive learning~\cite{khosla2021supervised} is proposed to extend the self-supervised batch contrastive approach to the fully supervised setting which can effectively leverage label information.

Based on the effectiveness of Contrastive Learning, the differences between this study and existing works are given below. First, we propose a Siamese Contrastive Embedding Network (SCEN) to excavate the discriminative prototypes of state and object, respectively. Besides, we present a State Transition Module (STM) to produce a virtual composition in training to improve the generalization of the proposed model. Second, we construct two contrastive spaces and utilize contrastive constraints to enforce the prototypes of state and object to be discriminative and generalized.

\begin{figure*}[htb]
	\centering
	\includegraphics[width=0.9\textwidth]{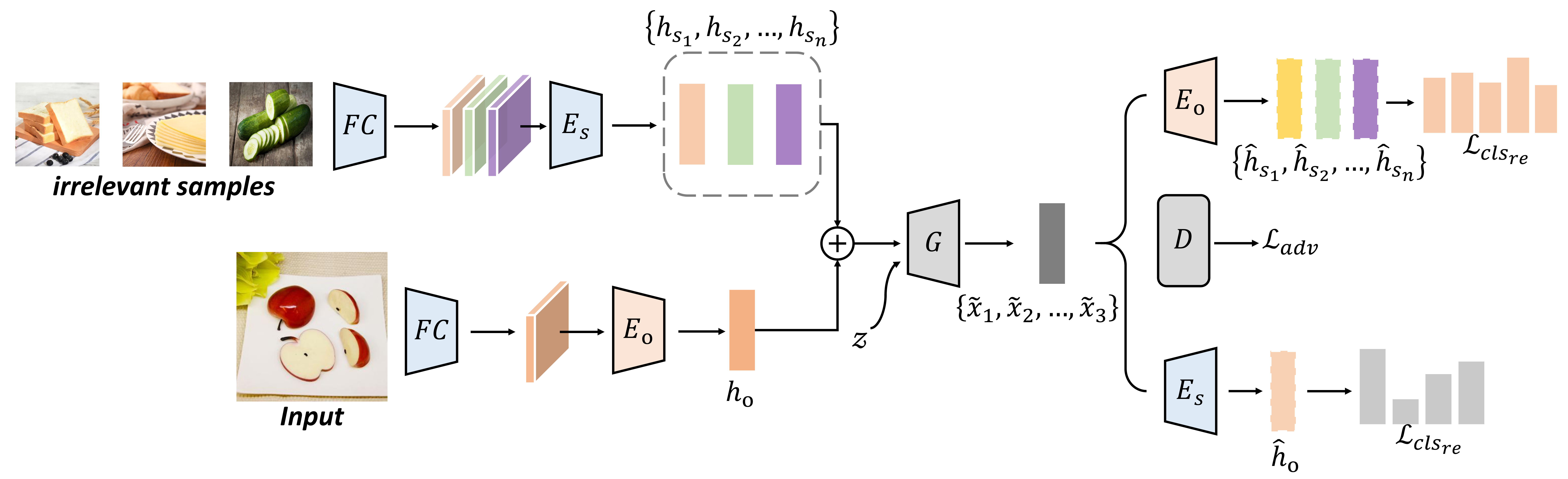} 
	\caption{The framework of our State Transition Module (STM). The proposed module aims to improve the generalization of State-Specific Encoder $E_{s}$ and Object-Specific Encoder $E_{o}$. To be specific, the Generator $G$ takes the state $h_{s}$ and object $h_{o}$ vectors as input and generates virtual samples, which is the input of Discriminator $D$. Although we do not gain the label for generated samples, we can utilize $E_{s}$ and $E_{o}$ to re-encode them and set classification losses as constraints.}
	\label{fig2}
\end{figure*}
\section{Approach}

The goal of CZSL is to recognize the novel compositional samples whose labels are composed of a state (e.g., \emph{old}) and an object (e.g., \emph{tiger}). This is particularly challenging since various states significantly change the visual appearance of an object, which hinders the performance of the classifiers.

We propose a novel formulation to tackle the problem, namely Siamese Contrastive Embedding Network (SCEN), which constructs two independent embedding spaces and utilizes contrastive losses to guide corresponding feature extractors to excavate discriminative prototypes of state and object separately. The overview of our approach is shown in~\cref{fig2}.

\subsection{Problem Definition}

In CZSL setting, each image consists of two primitive concepts. i.e., a state and an object. Given $\mathcal{A}$ and $\mathcal{O}$ as two sets of states and objects, we can compose a set of state-object pairs, i.e., $ \mathcal{C} = \mathcal{A} \times \mathcal{O} = \{ (a, o) \mid a \in \mathcal{A}, o \in \mathcal{O} \} $. Besides, we denote training set as $\mathcal{D}_{tr}=\{ (i,c) \mid i \in \mathcal{I}^{s}, c \in \mathcal{C}^{s} \} $, where $\mathcal{I}^{s}$ is the image set known in training, and $\mathcal{C}^{s}$ is a subset of $\mathcal{C}$ containing the corresponding labels. In the traditional Zero-Shot Learning setting, training and testing label are disjoint, i.e., $\mathcal{C}^{s} \cap \mathcal{C}^{u} = \varnothing$, where $\mathcal{C}^{s}$, $\mathcal{C}^{u}$ are two subsets of $\mathcal{C}$ seen/unseen in training. In this case, the model only needs to predict the compositions drawn from $C^{u}$ in testing~\cite{misra2017red}. In this paper, we follow the setting of Generalized ZSL~\cite{xian2018zero} where testing samples can be drawn from either seen or unseen compositions, i.e., $\mathcal{C}^{s} \cup \mathcal{C}^{u}$, which is more challenging on account of the larger prediction space and the dominant bias to seen compositions~\cite{purushwalkam2019task}. To sum up, CZSL aims to learn a mapping function $\mathcal{I} \to \mathcal{C}^{s} \cup \mathcal{C}^{u}$ that is trained on $\{\mathcal{I}^{s}, \mathcal{C}^{s}  \}$, in which $\mathcal{C}$ is composed of two primitive concepts drawn from $\mathcal{A}$ and $\mathcal{O}$.


\subsection{Siamese Contrastive Embedding Network}
Due to the entanglement between state and object into an image that influences the final classification, we design a Siamese Contrastive Embedding Network (SCEN) to better materialize the discriminative prototypes of state and object, which can effectively improve the accuracy of recognition models. The overall architecture is illustrated in~\cref{fig_fr}. The SCEN is composed of a feature extractor $FC$, a State-Specific Encoder $E_{s}$, a Object-Specific Encoder $E_{o}$, and a State Transition Module (STM).

\textbf{Specific database.} Let us consider a training sample, such as \emph{sliced apple} in~\cref{fig1}. As we all know, from our training set, that object \emph{apple} comes in various states such as \emph{caramelized} and state \emph{sliced} also does not just modify a single object, e.g., \emph{sliced banana}. Therefore, these sample points with overlapping information might have potential relationships. Based on this idea, we set up three specific databases $D_{s}$, $D_{o}$, and $D_{ir}$ to excavate discriminative state and object factors, respectively. $D_{s}$ is the set of compositions consisting of constant state and various objects, named State-constant database, while the Object-constant database $D_{o}$ is defined as the set of compositions made up of various states and a constant object. In addition, $D_{ir}$ is the set of compositions formed from various states and objects, which are both different from the state and object of input instances. For instance, given an image as input which consists of a state $\hat{a}$ and an object $\hat{o}$, i.e., $x = (\hat{a}, \hat{o}) \in \mathcal{I}^{s}$, the $D_{o}$ is denoted as:
\begin{equation}
\label{eq2}
\mathcal{D}_{o}= \{ (a, o) \mid o = \hat{o}, (a, o) \in \mathcal{C}^{s} \}.
\end{equation}
Analogously, the State-specific database $D_{s}$ is denoted as:
\begin{equation}
\label{eq3}
\mathcal{D}_{s}= \{ (a, o) \mid a = \hat{a}, (a, o) \in \mathcal{C}^{s} \}.
\end{equation}
And the irrelevant database $D_{ir}$ is denoted as:
\begin{equation}
\label{eq4}
\mathcal{D}_{ir}= \{ (a, o) \mid a \neq \hat{a}, o \neq \hat{o}, (a, o) \in \mathcal{C}^{s} \}.
\end{equation}

\textbf{Siamese Contrastive Space.} Based on sets of specific databases being set up, the visual feature $\boldsymbol{x}$, extracted by the feature extractor $FC$, are separately projected into two independent contrastive embedding spaces to extract prototypes of state $h_{s}$ and object $h_{o}$:
\begin{equation}
\begin{aligned}
\label{eq5}
& h_{s} = E_{s}(\boldsymbol{x}),\\
& h_{o} = E_{o}(\boldsymbol{x}).
\end{aligned}
\end{equation}

We hope that $h_{s}$ and $h_{o}$ contain information that is separately sensitive to classifiers for compositions recognition. Therefore, we aim to utilize contrastive learning as a constraint condition to extract discriminative prototypes of state and object. However, a general contrastive loss simultaneously cannot extract their discriminative representations due to their interaction. Based on this problem, we define state-based contrastive loss $\mathcal{L}_{scl}$ and object-based contrastive loss $\mathcal{L}_{ocl}$ as constraints to enforce the model to extract discriminative primitives.

To be specific, we set $h_{s}$ as an anchor in the state-based contrastive space. Meanwhile, $h_{s}^{ss}$ selected from $D_{s}$ is set as a positive point while $k$ negative samples selected from the $D_{ir}$ are denoted as $\{ h_{s_{1}}^{ir}, ..., h_{s_{k}}^{ir} \}$. We aim to decrease the distance between the anchor $h_{s}$ and the positive instance $h_{s}^{ss}$, increasing the distance between $h_{s}$ and each negative point $h_{s_{i}}^{ir}$ to extract discriminative prototype of state. Therefore, the state-specific loss function $\mathcal{L}_{scl}$ in the state-based contrastive space can be calculated as follows:
\begin{equation}
\label{eq6}
\mathcal{L}_{scl} = -\log \frac{exp((h_{s})^{\top} h_{s}^{ss}/\tau_{s})}{exp((h_{s})^{\top} h_{s}^{ss}/\tau_{s}) + \sum\limits_{i=1}^{K} {exp((h_{s})^{\top} h_{s_{i}}^{ir}/\tau_{s} )}},
\end{equation}
where $\tau_{s} > 0$ is the temperature parameter for the contrastive embedding and $K$ is the number of negative samples. It is obvious that the larger $K$ we set, the longer time the training process cost. The larger number of negative samples encourages State-Specific Encoders $E_{s}$ to excavate a more representative state prototype, which can be generalized to novel compositions.

As a Siamese Contrastive Embedding space, similar to the state-based contrastive space, we denote $h_{o}$ embedded from the same input visual features as an anchor in the object-based contrastive space, and $\{ h_{o_{1}}^{ir}, ..., h_{o_{k}}^{ir} \}$ as negative points embedded from irrelevant database $D_{ir}$. Therefore, the object-specific loss function $\mathcal{L}_{ocl}$ can be defined in the object contrastive space as:
\begin{equation}
\label{eq7}
\mathcal{L}_{ocl} = -\log \frac{exp((h_{o})^{\top} h_{o}^{os}/\tau_{o})}{exp((h_{o})^{\top} h_{o}^{os}/\tau_{o}) + \sum\limits_{j=1}^{K} {exp((h_{o})^{\top} h_{o_{j}}^{ir}/\tau_{o} )}},
\end{equation}
where $\tau_{o} > 0$ is the temperature parameter for the contrastive embedding and $K$ is the number of negative samples.

Considering that the prototypes of state and object should be optimized in the same direction, we share irrelevant samples for two contrastive spaces $\{ h_{o_{1}}^{ir}, ..., h_{o_{k}}^{ir} \}$ as negative points, which effectively avoids the problem of unbalanced optimization between $E_{s}$ and $E_{o}$.

Finally, we introduce classification losses to guide classifiers to recognize prototypes of state and object, respectively, which is formulated as:
\begin{equation}
\label{eq8}
\mathcal{L}_{cls} = C_{a} (h_{s}, a) + C_{o}(h_{o}, o),
\end{equation}
where $C_{a}$ and $C_{o}$ are both fully connected layers with the cross-entropy loss trained to classify state and object respectively. The prototypes of state and object can be further preserved in the composition with the supervision of classification losses. 

Thus, the total loss function in Siamese Contrastive Space $\mathcal{L}_{cts}$ can be formulated as:
\begin{equation}
\label{eq8}
\mathcal{L}_{cts} = \mathcal{L}_{scl} + \mathcal{L}_{ocl} + \mathcal{L}_{cls}.
\end{equation}

\textbf{State Transition Module.} In order to enforce the SCEN to be generalized to novel samples that do not appear in the training stage, we aim to produce virtual samples to augment the diversity of training compositions, alleviating the domain gap between training and testing data. Therefore, we propose a State Transition Module (STM), which consists of a State-Specific encoder $E_{s}$, a Object-Specific encoder $E_{o}$, a Generator $G$ and a Discriminator $D$~\cite{goodfellow2014generative}. The architecture is shown in~\cref{fig2}.

Let us consider two objects, namely \emph{apple} and \emph{banana}. As is known to all, from our training set, that \emph{apple} can be \emph{ripe} while \emph{banana} can be \emph{caramelized} since there appears at least one image in training. However, there exists \emph{caramelized apple} and \emph{ripe banana} compositions in the testing set while the training set does not. Therefore, we can conclude that the object has the possibility of forming a new combination with various states. Based on this discovery, we aim to utilize the Generator $G$ to produce virtual compositions with the input of various states and a given object. However, such generation with the random combination can produce many irrational compositions, which will actually widen the domain gap between seen and unseen data. For instance, \emph{cored banana} and \emph{squished apple} do not appear in the testing set or even exist in reality. Thus, we design a Discriminator $D$ to distinguish whether a composition is composed by the generator $G$.

In particular, the Generator $G$ takes as input the prototype of an object $h_{o}$ and another state $\tilde{h}_{s}$ to generate virtual compositions that never appear in training. Then, the Discriminator $D$ takes the real samples $x_{a, o}$ as input and determines which are produced by the Generator $G$. $G$ and $D$ can be optimized by the following adversarial objective:
\begin{equation}
\begin{aligned}
\label{eq9}
\max \limits_{D}\min \limits_{G, E_{s}, E_{o}} V(G,D) = & \mathbb{E}_{a, o} (\log D(x_{a, o})) +\\
& \mathbb{E}_{h_{\tilde{s}}, h_{o}}(\log(1 - D(G(h_{\tilde{s}}, h_{o})))),
\end{aligned}
\end{equation}
where $G(h_{\tilde{s}}, h_{o}) = \hat{x}_{\tilde{s}, o}$. $G$ tries to minimize $V(G, D)$ while $D$ tries to maximize it.

The goal of improving the $E_{s}$ and $E_{o}$  performance is to be generalized to novel compositions in testing, but the generated samples do not have labels as supervision. Thus, we re-encode the generated samples to extract prototypes of state and object again, and design a re-classification loss to constrain them, which is formulated as follows:
\begin{equation}
\label{eq10}
\mathcal{L}_{cls_{re}} = C_{a} (E_{s}(G(h_{\tilde{s}}, h_{o})), \tilde{a}) + C_{o}(E_{o}(G(h_{\tilde{s}}, h_{o})), o).
\end{equation}

The total loss function of State Transition Module $\mathcal{L}_{stm}$ is formulated as follows:
\begin{equation}
\label{11}
\mathcal{L}_{stm} = \max \limits_{D}\min \limits_{G, E_{s}, E_{o}} V(G,D) + \mathcal{L}_{cls_{re}}.
\end{equation}

Eventually, the final loss of our proposed framework is formulated as:
\begin{equation}
\label{11}
\mathcal{L}_{total} = \alpha \mathcal{L}_{cts} + \beta \mathcal{L}_{stm},
\end{equation}
where $\alpha$ and $\beta $ are the weighting coefficients to balance the influence of each loss function, respectively.


\subsection{Inference}
In the training stage, the model is trained to estimate the likelihood $p(I=i\mid A=a, O=o)$ for image $i$ conditioned on state $s$ and object $o$. The inference takes place in both $\mathcal{C}^{s}$ and $\mathcal{C}^{u}$. In the inference, model embeds an image as $\boldsymbol{x}$ and extract prototypes of state and object, i.e., $h_{a}$ and $h_{s}$ with trained $E_{s}$ and $E_{o}$, respectively. Then the state and object of the most similar prototypes are taken as the prediction. The inference rule can be parameterized as:
\begin{equation}
\label{eq1}
	c_{t} = (a_{t}, o_{t}) = \mathop{\arg\max}\limits_{(a, o) \in \mathcal{C}}p(i \mid E_{s}(\boldsymbol{x}), E_{o}(\boldsymbol{x})).
\end{equation}

\section{Experiment}

In this section, all datasets and evaluation protocols are introduced concretely. Then, we present the implementation details and the comparison of experimental results with other state-of-the-art methods. Eventually, ablation studies prove the effectiveness of the method we proposed.
\subsection{Experimental Setup}
\textbf{Datasets.} Our proposed method is evaluated on three CZSL benchmark datasets, i.e., MIT-States~\cite{isola2015discovering}, UT-Zappos~\cite{yu2014fine}, and C-GQA~\cite{naeem2021learning}.

MIT-States contains 53753 images, e.g., \emph{young cat} and \emph{rusty bike}, with 115 states and 245 objects in total. MIT-States has 1962 available compositions where 1262 state-object pairs are seen in the training stage, leaving 700 pairs unseen. UT-Zappos contains 50025 images of shoes, e.g., \emph{Cotton Sandals} and \emph{Suede Slippers}, with 16 states and 12 objects. In UT-Zappos, there are 116 state-object pairs, 83 pairs of which are used for training, while the other 33 pairs are unseen in training. As for C-GQA dataset, it contains over 9500 compositions that make it most extensive dataset for CZSL. The detailed information of each dataset is summarized in~\cref{tab1}.

\begin{table*}[h]
\centering
\begin{tabular}{ccc|cccccccc}
\hline
\multicolumn{3}{c|}{}                      & \multicolumn{2}{c}{\textbf{Training}} & \multicolumn{3}{c}{\textbf{Validation}} & \multicolumn{3}{c}{\textbf{Test}}      \\
\textbf{Dataset} & $s$ & $o$ & $c_{s}$        & $i$       & $c_{s}$  & $c_{u}$ & $i$ & $c_{s}$ & $c_{u}$ & $i$ \\ \hline
MIT-States~\cite{isola2015discovering}       & 115        & 245        & 1262               & 30k            & 300          & 300         & 10k      & 400         & 400         & 13k      \\
UT-Zappos~\cite{yu2014fine}        & 16         & 12         & 83                 & 23k            & 15           & 15          & 3k & 18          & 18          & 3000       \\
C-GQA~\cite{naeem2021learning}            & 453        & 870        & 6963               & 26k            & 1173         & 1368        & 7k & 1022        & 1047        & 5k       \\ \hline
\end{tabular}
\caption{Datasets used in our experiments and their statistics. We use three datasets to benchmark our method against the baseline states. states $s$, objects $o$, seen compositions $c_{s}$, unseen compositions $c_{u}$, images $i$.}
\label{tab1}
\end{table*}

\begin{table*}[htb]
\centering
\begin{tabular}{c|cccccccccccccc}
\hline
& \multicolumn{7}{c|}{\textbf{MIT-States}} & \multicolumn{7}{c}{\textbf{UT-Zappos}} \\
\multirow{2}{*}{\textbf{Method}}& \multicolumn{2}{c}{AUC} & \multicolumn{3}{c}{Best} & \multicolumn{2}{c|}{}&  \multicolumn{2}{c}{AUC} & \multicolumn{3}{c}{Best} & \multicolumn{2}{c}{} \\
& Val        & Test       & HM     & Seen  & Unseen  & s    & \multicolumn{1}{c|}{o}
& Val        & Test       & HM     & Seen   & Unseen  & s         & o        \\ \hline
AttOp~\cite{nagarajan2018attributes}                            & 2.5        & 1.6        & 9.9    & 14.3  & 17.4    & 21.1 & \multicolumn{1}{c|}{23.6} & 21.5       & 25.9       & 40.8   & 59.8   & 54.2   & 38.9      & 69.6     \\
LE+~\cite{misra2017red}                              & 3.0        & 2.0        & 10.7   & 15.0  & 20.1    & 23.5 & \multicolumn{1}{c|}{26.3} & 26.4       & 25.7       & 41.0   & 53.0   & 61.9   & 41.2      & 69.2     \\
TMN~\cite{purushwalkam2019task}                              & 3.5        & 2.9        & 13.0   & 20.2  & 20.1    & 23.3 & \multicolumn{1}{c|}{26.5} & 36.8       & 29.3       & 45.0   & 58.7   & 60.0   & 40.8      & 69.9     \\
SymNet~\cite{li2020symmetry}                           & 4.3        & 3.0        & 16.1   & 24.4  & 25.2    & 26.3 & \multicolumn{1}{c|}{28.3} & 25.9       & 23.9       & 39.2   & 53.3   & 57.9   & 40.5      & 71.2     \\
CGE~\cite{naeem2021learning}                              & 6.8        & 5.1        & 17.2   & 28.7  & \textbf{25.3}    & 27.9 & \multicolumn{1}{c|}{32.0} & 38.7       & 26.4       & 41.2   & 56.8   & \textbf{63.6}   & 45.0      & 73.9     \\
CompCos~\cite{mancini2021open}                          & 5.9        & 4.5        & 16.4   & 25.3  & 24.6    & 27.9 & \multicolumn{1}{c|}{31.8} & 38.6       & 28.7       & 43.1   & 59.8   & 62.5   & 44.7      & 73.5     \\ \hline
\textbf{Ours}       & \textbf{7.2}          & \textbf{5.3}          & \textbf{18.4}      & \textbf{29.9}     & 25.2       & \textbf{28.2}    & \multicolumn{1}{c|}{\textbf{32.2}}    & \textbf{40.2}       & \textbf{32.0}       & \textbf{47.8}   & \textbf{63.5}   & 63.1   & \textbf{47.3}      & \textbf{75.6}    \\ \hline
\end{tabular}
\caption{The state-of-the-art comparisons on UT-Zappos and MIT-States. \textit{U} and \textit{S} are the accuracies tested on unseen classes and seen classes in CZSL, respectively. \textit{H} is the harmonic mean value of \textit{U} and \textit{S}. The best results are marked in bold.}
\label{tab2}
\end{table*}

\begin{table}[htb]
\footnotesize
\centering
\begin{tabular}{c|ccccccc}
\hline
& \multicolumn{7}{c}{\textbf{C-GQA}} \\
\multirow{2}{*}{\textbf{Method}}
& \multicolumn{2}{c}{AUC} & \multicolumn{3}{c}{Best} &  \\
& Val        & Test       & HM     & Seen  & Unseen  & s    & o   \\ \hline
AttOp~\cite{nagarajan2018attributes}                            & 0.9        & 0.3        & 2.9    & 11.8  & 3.9    & 8.3 & 12.5    \\
LE+~\cite{misra2017red}                              & 1.2        & 0.6        & 5.3   & 16.1  & 5.0    & 7.4 & 15.6     \\
TMN~\cite{purushwalkam2019task}                              & 2.2        & 1.1        & 7.7   & 21.6  & 6.3    & 9.7 & 20.5     \\
SymNet~\cite{li2020symmetry}                           & 3.3        & 1.8        & 9.8   & 25.2  & 9.2   & 14.5 & 20.2    \\
CGE~\cite{naeem2021learning}                              & 3.6        & 2.5        & 11.9   & 27.5  & 11.7    & 12.7 & 26.9    \\ \hline
\textbf{Ours}                    & \textbf{4.0}       &\textbf{2.9}        &\textbf{12.4}   & \textbf{28.9} &\textbf{12.1}   &\textbf{13.6} &\textbf{27.9} \\ \hline
\end{tabular}
\caption{The state-of-the-art comparisons on recent proposed C-GQA dataset.}
\label{tab3}
\end{table}

\begin{table*}[htb]
\centering
\begin{tabular}{c|cccccccccccccc}
\hline
                                 & \multicolumn{7}{c|}{\textbf{MIT-States}}                                               & \multicolumn{7}{c}{\textbf{UT-Zappos}}                                    \\
\multirow{2}{*}{\textbf{Method}} & \multicolumn{2}{c}{AUC} & \multicolumn{3}{c}{Best} & \multicolumn{2}{c|}{}            & \multicolumn{2}{c}{AUC} & \multicolumn{3}{c}{Best} & \multicolumn{2}{c}{} \\
                                 & Val        & Test       & HM     & Seen  & Unseen  & s    & \multicolumn{1}{c|}{o}    & Val        & Test       & HM     & Seen   & Useen  & s         & o        \\ \hline
Base                             & 2.5        & 1.6        & 9.9    & 14.3  & 17.4    & 21.1 & \multicolumn{1}{c|}{23.6} & 21.5       & 25.9       & 40.8   & 59.8   & 54.2   & 38.9      & 69.6     \\
+$\mathcal{L}_{cts}$                               & 5.8        & 4.3        & 16.5   & 25.3  & 24.7    & 27.9 & \multicolumn{1}{c|}{31.6} & 32.6       & 26.4       & 44.2   & 60.0   & 61.2   & 42.5      & 71.3     \\
+$\mathcal{L}_{stm}$                               &  4.8         & 3.9          & 13.3      & 22.5     & 21.9       & 27.6    & \multicolumn{1}{c|}{30.2}    & 34.6          & 28.5          & 43.9      & 59.8      & 61.1      & 46.9         & 72.1        \\ \hline
$\mathcal{L}_{cts}$+$\mathcal{L}_{stm}$                    & \textbf{7.2}          & \textbf{5.3}          & \textbf{18.4}      & \textbf{29.2}     & \textbf{25.2}       & \textbf{28.2}    & \multicolumn{1}{c|}{\textbf{32.2}}    & \textbf{39.0}          & \textbf{29.7}          & \textbf{47.8}      & \textbf{63.3}      & \textbf{62.5}      & \textbf{47.3}         & \textbf{74.4}    \\ \hline   
\end{tabular}
\caption{Ablation studies for Compositional Zero-Shot Learning on MIT-States and UT-Zappos datasets.}
\label{tab4}
\end{table*}

\textbf{Evaluation Metrics.} We evaluate the performance according to prediction accuracy for recognizing seen and unseen compositions. Following the setting of~\cite{purushwalkam2019task}, we compute the accuracy in two situations: 1) \emph{Seen}, testing only on seen compositions; 2) \emph{Unseen}, testing only on unseen compositions. Based on these, we can compute \emph{Harmonic Mean} $\emph{HM}$ of the two metrics, which balances the performance between seen and unseen accuracies. Eventually, we compute 4) \emph{Area Under the Curve (AUC)} to quantify the overall performance of both seen and unseen accuracy at different operating points with respect to the bias. Following~\cite{purushwalkam2019task, chao2016empirical}, we utilize a calibration bias to trade off between the prediction scores of seen and unseen pairs. As the calibration bias varies, we can draw a seen-unseen accuracy curve where the AUC metric can be computed. 

\textbf{Implementation Details.} For each image, we extract a 1024 dimensional visual feature vector using ResNet-18~\cite{he2016deep} pre-trained on the ImageNet dataset~\cite{russakovsky2015imagenet}. We separately extract a 300-dimensional feature vectors for both states and objects with $E_{s}$ and $E_{o}$, which is implemented with two fully-connected layers and ReLU activation. Our model is implemented with PyTorch~\cite{paszke2019pytorch} and optimized by ADAM optimizer~\cite{kingma2014adam} on an NVIDIA GTX 1080Ti GPU. In addition, we set the learning rate as 0.00004, batch size as 128, and  the number of negative samples $K$ as 10. For the MIT-States dataset, the training time is approximately 3 hours for 800 epochs. For the UT-Zappos dataset, it takes around 1 hour for 500 epochs in training. As for C-GQA, it spends around 4 hours for 1000 epochs in training.

\subsection{Comparison with State-of-the-Arts}
We compare our experiments with the state-of-the-art in~\cref{tab2} and show that our Siamese Contrastive Embedding Network (SCEN) outperforms all previous methods in three benchmark datasets, which includes recent proposed C-GQA dataset~\cite{naeem2021learning}. Our detailed observations are as follows.


\textbf{Generalized CZSL performance.} For the CZSL task, our SCEN achieves a test AUC of 5.3\%, which achieves the best result on MIT-States. In addition, our method significantly boosts the state-of-the-art harmonic mean, i.e., 17.2\% to 18.4\%. When it refers to state and object prediction accuracy, we can observe an improvement from 27.9\% to 28.2 \% for states and 31.8\% to 32.2\% for objects.

Similar observations are confirmed on UT-Zappos, in which we can achieve a superior improvement on state-of-the-arts with an AUC of 32.0\% compared to 28.7\% from Compcos. In addition, our proposed model performs the best harmonic mean 47.8\% and improves around 4.5\% compared with the Compcos.

Finally, on the recent proposed splits of the C-GQA dataset, which is shown in~\cref{tab3}, we also achieve the best test AUC of 4.0\%. Since C-GQA is a large number of compositions (over 9.3k concepts), which is more complex than MIT-States and UT-Zappos for recognition. The state and the object accuracies of our method are 13.6\% and 27.9\%, which are both higher than state-of-the-arts. In addition, our best seen and unseen accuracies (28.9\% and 12.1\%) also achieve the best results on this new dataset.

According to the signficant improvement on three challenging datasets, we can conclude that our proposed Siamese Contrastive Embedding Network (SCEN) can not only effectively extract discriminative prototypes of state and object, but also improve the robust of the model for unseen compositions recognition.

\subsection{Ablation Study}

We now make an ablation study to evaluate the effectiveness of the Siamese Contrastive Embedding Network. We take a single classification model as a base model, which trains two classifiers to recognize states and objects separately. Meanwhile, we train three variants by adding Siamese Contrastive space, adding State Transition Module (STM), or adding both of them, which is denoted as \emph{base model}, \emph{base model} + $\mathcal{L}_{cts}$, and \emph{base model} + $\mathcal{L}_{cts}$ + $\mathcal{L}_{stm}$, respectively. According to the showing experiment of each setting as shown in~\cref{tab4}, every variant tends to be more superior performance than the \emph{base model}. The combination of two variants we proposed achieves the best improvement, which demonstrates that different components promote each other and work together to improve the performance of SCEN significantly. In addition, the result of the \emph{base model} with $\mathcal{L}_{cts}$ proves that our proposed model successfully excavates discriminative prototypes of states and objects, which is better for compositions recognition. Meanwhile, the significant improvement after adding $\mathcal{L}_{stm}$ indicates the necessity of the proposed State Transition Module, which can effectively alleviate the domain gap between training and testing data.  Finally, the result of the \emph{base model} with adding $\mathcal{L}_{cts}$ and $\mathcal{L}_{stm}$ achieves the best, which shows that they improve the performance of the model together, not affected by each other.


\subsection{Qualitative Results}
\begin{figure*}[t]
\centering
\includegraphics[width=0.85\textwidth]{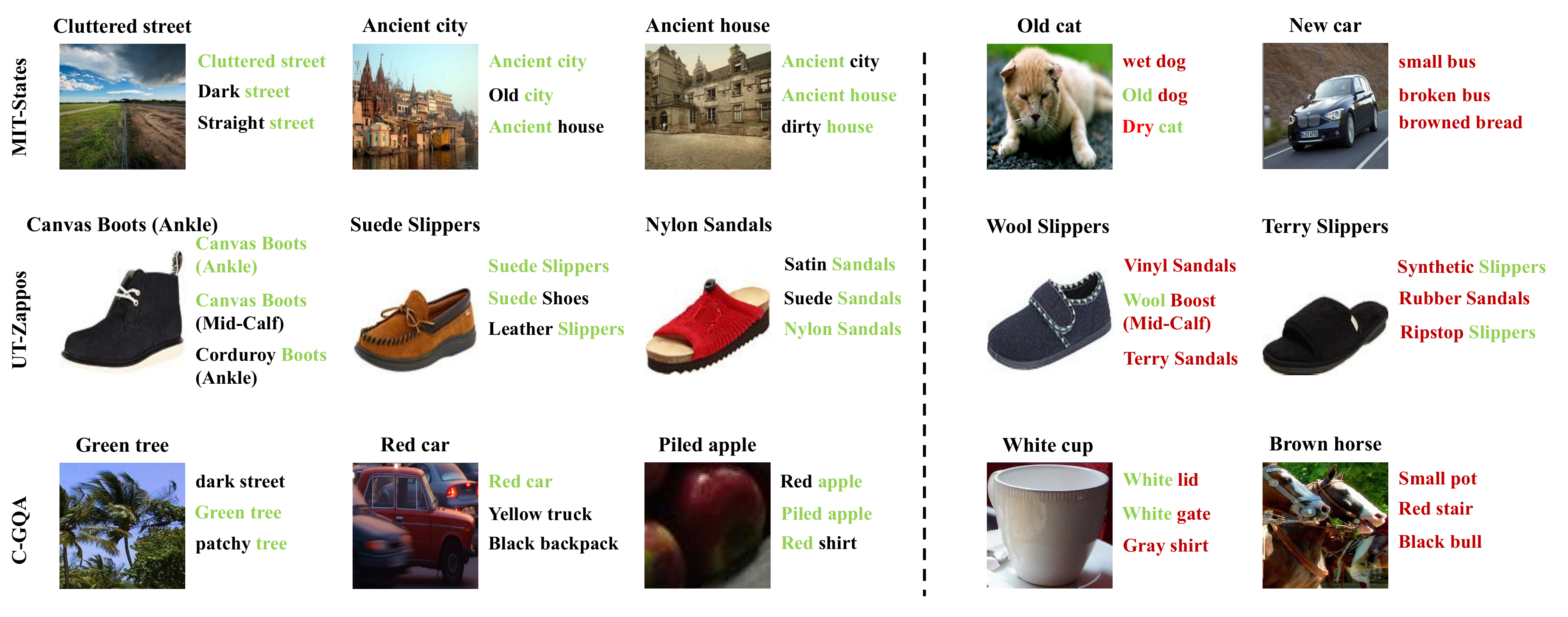} 
\caption{Qualitative results. We show the top-3 predictions of our proposed model for some instances. From the first three columns, we can observe that all predictions of the model indicate it can categorize accurately. However, the model is only incentivized when it matches the label. The task of CZSL is a multi-label one, and future datasets need to account for this. The last two columns show some examples of suboptimal labels and wrong predictions.}
\label{fig4}
\end{figure*}

We show some qualitative results for the novel compositions with top-3 predictions in~\cref{fig4}. The first three columns present some examples where the top prediction matches the label. For MIT-States and UT-Zappos datasets, we notice that the remaining two answers of the model can fundamentally capture at least one factor, which proves that the superior performance of our method. As for more complex C-GQA dataset, our model can give the correct answer in top-3 predictions, which shows the robustness of our proposed framework.

Meanwhile, the model can predict more combinations of unseen compositions, rather than being limited to that of seen compositions, which effectively alleviate the domain gap between seen and unseen samples.

In addition, the last two columns show the wrong prediction. For instance, in column 4 and row 2, the image of the \emph{Slippers} is misclassified as \emph{Sandals} or \emph{Boost}. This is because there exists a large number of training compositions so that the negative sample dataset may not contain the entire negative samples, such as \emph{Sandals} and \emph{Boost}, thus the model does not pay more attention to these pairs that are not included in $D_{ir}$. Besides, limited by the compositional class accuracy dependent on the number of groups associated with an object in the label space, the model may only focus on the state of the object in a certain aspect and ignore the state of the object labeled by the tag. For example, in column 4, row 1 the image of the \emph{cat} consists of texture and age both present in the label space of the dataset and the output of the model. However the label for this image only contains its age.

\begin{figure}[htb]
\centering
\includegraphics[width=0.45\textwidth]{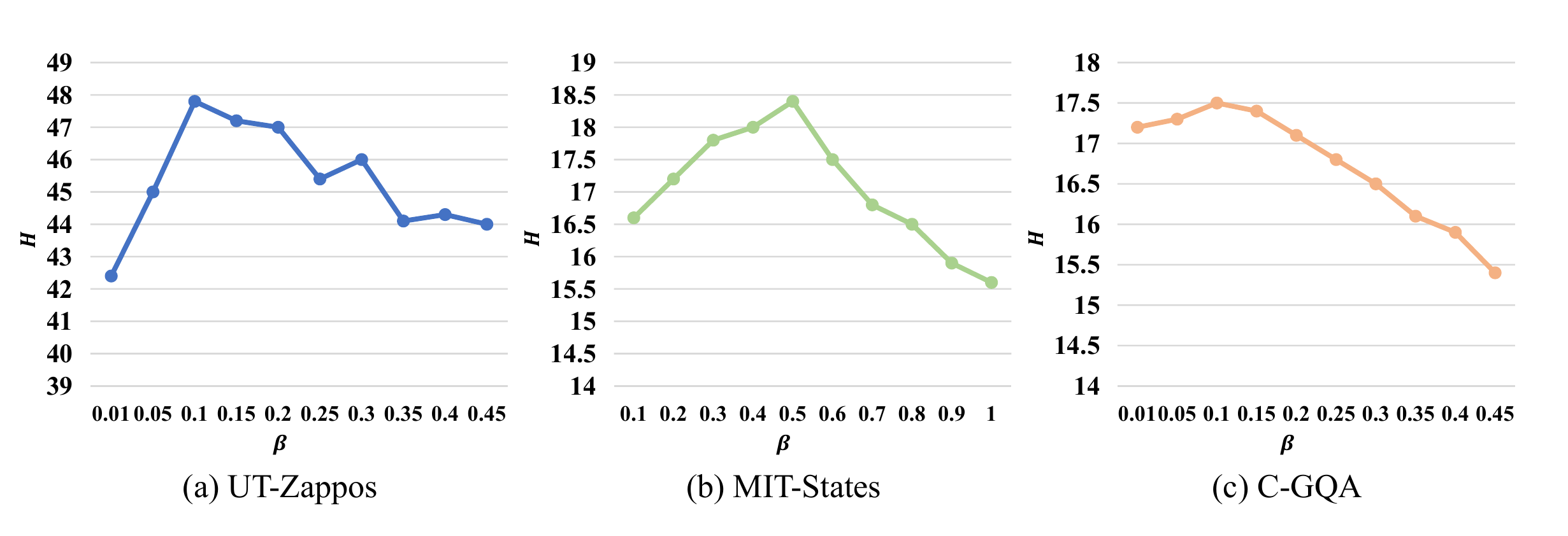} 
\caption{The influence of the weighting coefficient $\beta$ for \emph{H} value.}
\label{fig5}
\end{figure}

\begin{figure}[htb]
\centering
\includegraphics[width=0.45\textwidth]{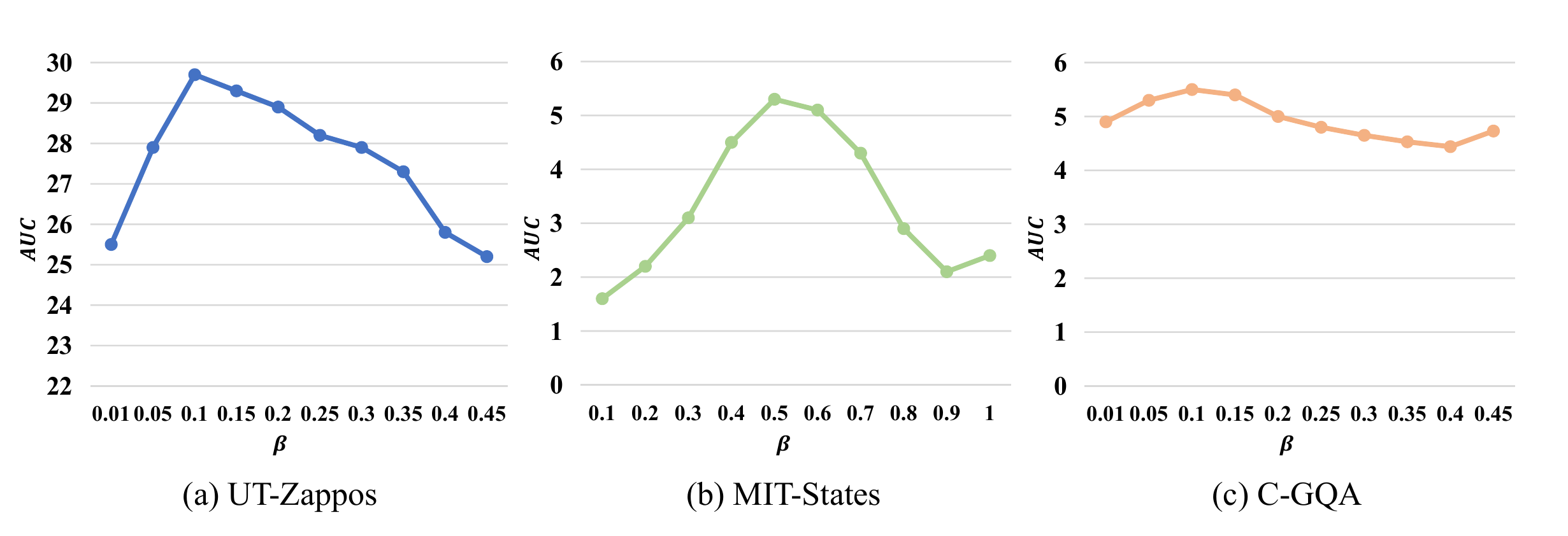} 
\caption{The influence of the weighting coefficient $\beta$ for \emph{AUC}.}
\label{fig6}
\end{figure}

\subsection{Hyper-Parameter Analysis.} We perform an experiment to demonstrate the effect of the weighting co-efficients $\alpha$ and $\beta$ for the loss functions $\mathcal{L}_{cts}$ and $\mathcal{L}_{stm}$ in our proposed model. As is shown in~\cref{fig5} and~\cref{fig6}. With the different $\alpha$ and $\beta$ setting, the \emph{Harmonic Mean} $\emph{HM}$ and $\emph{AUC}$ have a certain degree of change, which indicates that $\mathcal{L}_{cts}$ and $\mathcal{L}_{stm}$ dominate the performance of the entire model. Based on this situation, we set and fix $\alpha = 0.1$, changing the value of $\beta$ to observe the performance on different datasets. Finally, on MIT-States, UT-Zappos, and C-GQA datasets, we set $\beta = 0.5$, $\beta = 0.1$, and $\beta = 0.1$, which can achieve the best results, respectively.

\section{Conclusion}

In this paper, we propose a novel Siamese Contrastive Embedding Network (SCEN) to excavate discriminative prototypes of state and object for the CZSL task. We firstly project the visual feature into two contrastive spaces, where we set up \emph{state-constant} and \emph{object-constant} databases. Meanwhile, we design state-specific and object-specific loss functions as constraints, forcing them to contain discriminative corresponding information. In addition, we design a State Transition Module (STM) to produce virtual but rational compositions that never appear in training, which effectively augment the diversity of training data. The proposed module can provide a robust model that can excavate prototypes for seen samples and be generalized to novel compositions, where linear softmax classifiers can be trained to recognize compositions from both seen and unseen instances. The comparison and ablation study experiments demonstrate that our proposed CZSL framework has achieved state-of-the-arts on three challenging datasets.  

\section{Acknowledgements}

Our work was supported in part by the National Natural Science Foundation of China under Grant 62132016, Grant 62171343, Grant 62071361, in part by Key Research and Development Program of Shaanxi under Grant 2021ZDLGY01-03, and in part by the Fundamental Research Funds for the Central Universities ZDRC2102.


{\small
\bibliographystyle{ieee_fullname}
\bibliography{05950}
}

\end{document}